\documentclass[conference]{IEEEtran}
\IEEEoverridecommandlockouts
\usepackage{cite}
\usepackage{amsmath,amssymb,amsfonts}
\usepackage{graphicx}
\usepackage{textcomp}
\usepackage{xcolor}
\usepackage{url}
\usepackage{booktabs}
\usepackage{makecell}
\usepackage{hyperref}

\def\BibTeX{{\rm B\kern-.05em{\sc i\kern-.025em b}\kern-.08em
    T\kern-.1667em\lower.7ex\hbox{E}\kern-.125emX}}
\hypersetup{hidelinks}
\begin{document}

\title{Entropy-Punctured Bloom Filters for Memory-Efficient Machine Learning}

\author{
\IEEEauthorblockN{John Cartmell, Mihaela Cardei, Ionut Cardei}
\IEEEauthorblockA{
Department of Electrical Engineering and Computer Science\\
Florida Atlantic University\\
Boca Raton, FL, USA\\
jcartmell2023@fau.edu, mcardei@fau.edu, icardei@fau.edu\\
ORCID: 0000-0002-7014-4005, 0000-0003-2359-6196, 0009-0000-1050-768X
}
}

\maketitle

\begin{abstract}
Memory-efficient feature representations are increasingly important in machine learning settings where storage, transmission cost, bandwidth, or privacy constraints limit access to raw data. Bloom Filter (BF) encodings provide compact probabilistic representations of engineered features, but their behavior under structural compression and their applicability to regression tasks remain underexplored.

In this work, we propose entropy-punctured Bloom Filters, a memory-aware encoding strategy that removes low-variability bit positions identified using empirical entropy. Starting from fixed-length BF encodings of quantized features, the proposed approach produces reduced representations that preserve predictive structure while improving predictive efficiency relative to encoded representation size.

We evaluate the approach on diverse regression datasets, comparing raw features, Principal Component Analysis (PCA), Random Projection (RP), and Bloom Filter variants under leakage-free evaluation protocols and approximately matched representation sizes. Performance is assessed using ridge regression, XGBoost, and neural networks, with predictive efficiency measured as $R^2$ relative to encoded representation size per sample.

Results show that Bloom Filter encodings remain competitive with classical compressed representations while achieving substantial storage savings. Entropy-based puncturing further reduces representation size with minimal loss in predictive fidelity, yielding improved predictive efficiency. These findings demonstrate that entropy-punctured Bloom Filters provide an effective representation-level compression approach for memory-constrained machine learning.
\end{abstract}

%%%%%%%%%%%%%%%%%%%%%%%%%%%


%%%%%%%%%%%%%%%%%%%%%%%%%%%

\section{Introduction}
Bloom Filters (BFs) provide compact probabilistic set representations and have been widely used in database and network systems \cite{bloom1970}. In machine learning, they have been applied both as membership filters within pipelines \cite{mitzenmacher2018learned} and as feature encoders that transform raw attributes into compressed binary representations \cite{cartmell2025bfencoding}. Prior work has focused primarily on classification, demonstrating that Bloom Filter representations can reduce memory usage while maintaining competitive predictive performance.

However, the use of Bloom Filter encodings for regression under explicit memory constraints remains largely unexplored. Regression tasks require preserving continuous signal relationships and may therefore be more sensitive to distortion introduced by hashing collisions and quantization. At the same time, many practical machine learning systems operate under storage, bandwidth, or privacy constraints that limit access to full raw feature representations.

Classical dimensionality reduction methods such as Principal Component Analysis (PCA) \cite{jolliffe2002pca}, Random Projection (RP) \cite{johnson1984extensions}, and feature hashing \cite{weinberger2009feature} are typically evaluated primarily in terms of predictive accuracy. Under constrained-memory settings, predictive accuracy alone does not fully capture the trade-off between predictive performance and representation size.

Motivated by this setting, we introduce \emph{predictive efficiency}, defined as the coefficient of determination ($R^2$) relative to encoded representation size per sample, capturing predictive signal retained per unit storage.

We propose entropy-punctured Bloom Filters, a structural compression method that removes low-variability bit positions based on empirical entropy. Starting from fixed-length Bloom Filter encodings of tokenized and quantized features, the proposed approach reduces redundancy while preserving predictive structure. Bloom Filter bits that are almost always active or inactive exhibit low marginal variability across samples. Entropy-guided puncturing preferentially removes these near-constant positions while retaining positions that vary more substantially across samples.

Inspired by puncturing in communication systems~\cite{hagenauer1988rate,lin2004error}, we investigate whether selective bit removal can reduce encoded representation size while preserving predictive fidelity.

We evaluate the proposed approach across tabular, time-series, and text regression datasets using ridge regression \cite{hoerl1970ridge}, XGBoost \cite{chen2016xgboost}, and neural networks \cite{goodfellow2016deep}. Comparisons against raw features and classical compression methods are performed under leakage-free evaluation protocols and approximately matched representation sizes. Results show that entropy-punctured Bloom Filters maintain predictive fidelity while substantially improving predictive efficiency.

The main contributions of this work are: (i) entropy-punctured Bloom Filters for memory-efficient machine learning, (ii) a storage-normalized evaluation of predictive fidelity relative to encoded representation size, and (iii) a controlled structural analysis of how entropy-ranked bit removal and Bloom Filter folding affect predictive fidelity and representation size.

%%%%%%%%%%%%%%%%%%%%%%%%%%%

\section{Related Work}
Bloom Filters (BFs) are widely used for compact set representation and membership testing~\cite{bloom1970}, with applications in databases, networking, and distributed systems. In machine learning, they have primarily been used as pre- or post-processing components that reduce computational or memory cost rather than as direct feature encodings~\cite{mitzenmacher2018learned}.

Classical dimensionality reduction methods such as Principal Component Analysis (PCA) and Random Projection (RP) compress data through linear transformations that preserve variance or approximate geometric structure~\cite{jolliffe2002pca,johnson1984extensions}. Feature hashing similarly produces fixed-length representations using hash functions for scalable handling of high-dimensional or sparse features~\cite{weinberger2009feature}. Unlike PCA and RP, hash-based approaches—including Bloom Filters—trade controlled collisions for compactness and are typically evaluated primarily in terms of predictive fidelity.

Prior work has explored Bloom Filter encodings for predictive modeling, with much of the existing literature emphasizing classification. Tokenized features can be compressed into compact binary representations while maintaining competitive classification performance~\cite{cartmell2025bfencoding}. Zaman et al.~\cite{zaman2024privacy} subsequently combined Bloom Filter encoding with local differential privacy for IoT data analysis and evaluated the resulting representations in downstream deep-learning tasks, including time-series forecasting. Their work principally examines the privacy-utility trade-off of perturbed Bloom Filter encodings. In contrast, the present work focuses on representation-level compression for regression through the selective removal of Bloom Filter positions under explicit memory constraints. Thus, although Bloom Filter-encoded inputs have previously been used for forecasting, their behavior under entropy-guided structural compression has received comparatively limited attention.

Our approach is related to information-theoretic notions of redundancy and efficiency. Marginal entropy quantifies the variability of individual Bloom Filter positions but does not measure their mutual information with the prediction target~\cite{shannon1948mathematical}. We therefore use entropy as a label-independent indicator of nearly constant bit positions rather than as a guarantee of predictive relevance. Puncturing in coding theory removes selected bits to increase coding rate with limited performance degradation~\cite{hagenauer1988rate,lin2004error,richardson2008modern}. Inspired by these ideas, we investigate whether removing low-entropy Bloom Filter positions can reduce representation size while preserving predictive fidelity.

More broadly, memory-aware and privacy-preserving learning has emphasized reducing resource usage in constrained environments~\cite{cartmell2026surveyprivacy}, though much of this work focuses on model compression rather than feature representation. In contrast, we study representation-level compression and use predictive efficiency as a storage-normalized summary of predictive performance under memory constraints.

%%%%%%%%%%%%%%%%%%%%%%%%%%%

\section{Methods}
\subsection{Bloom Filter Feature Encoding}
Each sample is transformed into a set of discrete feature symbols and encoded into a fixed-length Bloom Filter (BF) representation. Numeric features are discretized into bins, categorical features are used directly, and text inputs are tokenized. For time-series datasets, lagged features are constructed prior to discretization. Continuous-valued features are processed directly through discretization without additional normalization, allowing the encoding procedure to operate consistently across heterogeneous feature types.

Each symbol encodes both feature identity and value (e.g., feature name together with a quantized bin index or token). Symbols are mapped into an $m$-bit vector using $k$-hashing operations, producing a binary representation for a dataset containing $n$ samples:
\[
BF \in \{0,1\}^{n \times m}.
\]

For temporal datasets containing multiple lag windows, separate Bloom Filter encodings may be constructed for different temporal segments and concatenated into a single representation. This enables local temporal structure to be captured while maintaining a fixed-length representation.

A key property of this encoding is that dimensionality remains fixed regardless of the number of generated symbols, enabling flexible feature engineering under constant storage cost. Additional implementation details and encoding examples are provided in prior work on Bloom Filter feature encodings for classification~\cite{cartmell2025bfencoding}.

\subsection{Entropy-Based Puncturing}
To reduce representational redundancy, we apply entropy-guided puncturing to the Bloom Filter representation, drawing on the general concept of puncturing from coding theory~\cite{hagenauer1988rate}. Let $p_j$ denote the empirical activation probability of bit position $j$ across the training samples. The marginal entropy $H_j$ of each bit is~\cite{shannon1948mathematical}
\begin{equation}
H_j = -\left[p_j\log_2p_j+(1-p_j)\log_2(1-p_j)\right],
\end{equation}
where $0\log_2 0$ is defined as zero.

Given an entropy threshold $\tau$, the retained set of bit positions and its cardinality are
\begin{equation}
S_\tau =
\left\{j \in \{1,\ldots,m\}: H_j \ge \tau\right\},
\qquad
m_\tau = |S_\tau| \le m.
\end{equation}
The punctured representation consists only of the positions in $S_\tau$.

Each Bloom Filter position can be interpreted as a Bernoulli binary feature. Its variance is
\begin{equation}
\operatorname{Var}(B_j)=p_j(1-p_j).
\end{equation}
Binary entropy and variance are both symmetric about $p_j=0.5$ and increase monotonically as $p_j$ moves from $0$ toward $0.5$. Consequently, they induce the same ranking of binary bit positions, and variance-based pruning would select the same positions as entropy-based pruning for a fixed retention count.

Importantly, $H_j$ is a marginal quantity and does not measure the mutual information between bit $B_j$ and the prediction target. A low-entropy bit may still identify a rare but predictive event, while a high-entropy bit may vary without being predictive. Entropy is therefore used here as a label-independent indicator of near-constant and potentially redundant positions, rather than as a guarantee of predictive relevance. The controlled low-entropy, high-entropy, and random-removal experiments in Section~V empirically evaluate whether this structural criterion preserves predictive fidelity.

Entropy statistics are computed exclusively from the training portion of each evaluation fold. The resulting set $\mathcal{S}_{\tau}$ is then applied unchanged to the corresponding validation data, preventing information leakage.

\subsection{Parameters}
The encoding process is governed by four parameters:
\begin{itemize}
\item Bloom Filter size $m$;
\item number of hash functions $k$;
\item number of quantization bins $b$; and
\item entropy threshold $\tau$.
\end{itemize}

These parameters jointly control representation capacity, collision behavior, sparsity, and compression. They were selected using a lightweight heuristic intended to identify stable operating regimes without exhaustive parameter sweeps. The heuristic selects candidate Bloom Filter sizes based on dataset dimensionality and target representation size, chooses hash counts that maintain moderate Bloom Filter occupancy, and evaluates a limited range of entropy thresholds to balance predictive fidelity and compression efficiency.

Within each evaluation fold, the heuristic and entropy mask are determined exclusively from the training partition; the corresponding validation partition is used only for out-of-sample evaluation. All representations are evaluated using the same data splits and downstream model configurations. The resulting configurations provide practical trade-offs among collision rate, representation sparsity, and predictive efficiency while substantially reducing search complexity relative to exhaustive tuning.

\subsection{Structural Analysis}
To better understand how predictive signal is distributed within Bloom Filter representations, we evaluate several structural transformations and perturbations.

\textbf{Bit Removal:} In addition to entropy-based puncturing, we compare against random bit removal and high-entropy bit removal to assess where predictive signal is concentrated within the representation.

\textbf{Permutation:} Bit positions are randomly permuted prior to training to evaluate whether predictive performance depends on positional structure or on aggregate activation patterns.

\textbf{Folding:} The Bloom Filter is compressed by combining pairs of bit positions using logical OR operations leading to a BF size reduction to $m/2$ bits:
\begin{equation}
b'_i = b_i \lor b_{i + m/2}.
\end{equation}

This operation reduces the Bloom Filter size by approximately one-half while increasing collision rates through superposition of bit positions.

Together, these experiments evaluate the robustness of Bloom Filter representations under controlled structural perturbations and compression.

%%%%%%%%%%%%%%%%%%%%%%%%%%%

\section{Experiments}
\label{sec:experiments}
The experimental evaluation investigates both predictive fidelity and representation efficiency of Bloom Filter encodings and entropy-punctured Bloom Filters. Experiments compare raw feature representations with classical dimensionality reduction methods and Bloom Filter encodings across multiple datasets and regression model families.

The study is designed to evaluate three key properties of Bloom Filter encodings: predictive fidelity, predictive efficiency, and structural robustness.

\subsection{Datasets}
We evaluate the proposed approach across tabular, time-series, and text regression datasets. Table~\ref{tab:datasets} summarizes dataset characteristics including sample size, dimensionality, data modality, and raw memory footprint per sample.

The evaluation includes standard tabular regression datasets (Boston Housing~\cite{boston_dataset}, California Housing~\cite{california_dataset}, Abalone~\cite{abalone_dataset}, and Allstate Claims Severity~\cite{allstate_dataset}), temporal datasets with lagged feature construction (Airline Passengers~\cite{airline_dataset} and Bike Sharing~\cite{bike_dataset}), biomedical sensor data (Parkinson’s Telemonitoring~\cite{parkinsons_dataset}), and large-scale sparse text regression (Yelp Reviews~\cite{yelp_dataset}). Collectively, these datasets provide a heterogeneous testbed for evaluating predictive fidelity, compression efficiency, and structural robustness across different data modalities.

\begin{table}[t]
\centering
\caption{Summary of Datasets}
\begin{tabular}{@{}l l l l l@{}}
\toprule
Dataset & Type & Samples & Features & \makecell{Raw Size\\(bytes per sample)} \\
\midrule
Abalone & Tabular & 4,177 & 8 & 64 \\
Airline & Time series & 144 & 12 & 96 \\
Allstate & Tabular & 188,318 & 130 & 170 \\
Bike & Time series & 17,379 & 24 & 192 \\
Boston & Tabular & 506 & 13 & 104 \\
California & Tabular & 20,640 & 8 & 64 \\
Parkinson's & Tabular & 5,875 & 16 & 144 \\
Yelp & Text & 50,000 & Text & 512 \\
\bottomrule
\end{tabular}
\label{tab:datasets}
\end{table}

\subsection{Evaluation Protocol}
To ensure fair comparisons and prevent information leakage, the same training and held-out partitions are used for all representations within each evaluation split. All representation-specific preprocessing is fitted exclusively on the training partition and subsequently applied unchanged to the corresponding held-out data. This includes quantization, Bloom Filter parameter selection and encoding, entropy estimation and puncturing-mask construction, and the fitting of PCA and Random Projection transformations. No held-out observations are used to select encoding parameters or punctured bit positions.

Unless otherwise specified, model performance is estimated using 5-fold cross-validation~\cite{james2013isl}. The same folds are reused across representations, and the model configuration for a given learner is held fixed across those representations. Performance metrics are averaged across folds to provide comparable estimates of generalization performance.

The time-series datasets use order-preserving evaluation procedures. For the Bike Sharing dataset, a chronological 50-50 train-test split is used. For the Airline Passengers dataset, forward time-series splits ensure that all training observations precede the corresponding test observations.

Fixed random seeds are used for model initialization, data shuffling, dimensionality-reduction methods, and $k$-hashing operations where applicable, ensuring that the reported comparisons are reproducible.

\subsection{Bloom Filter Configuration}
Bloom Filter parameters were selected using the lightweight heuristic described in Section III. The heuristic determines filter size $m$, number of hash functions $k$, and entropy threshold $\tau$ based on dataset characteristics including token count, representation sparsity, and target encoded size. The objective is to balance Bloom Filter occupancy, collision behavior, compression efficiency, and predictive fidelity while avoiding exhaustive parameter sweeps.

For the Bike Sharing and Airline datasets, multiple Bloom Filters are constructed over different temporal windows and concatenated into a single representation. The reported parameters correspond to representative values across the component filters.

Overall, predictive performance remained relatively stable across moderate variations in parameter selection, as demonstrated in the sensitivity analysis experiments.

Table~\ref{tab:bf_config} summarizes the final dataset-specific configurations used in the experiments.

\begin{table}[t]
\centering
\caption{Bloom Filter Encoding Parameters Used for Each Dataset}
\label{tab:bf_config}
\begin{tabular}{lccc}
\toprule
Dataset & $m$ Bytes & $k$ Hashes & $\tau$ Threshold \\
\midrule
Abalone & 32 & 3 & 0.30 \\
Airline & 192 & 3 & 0.05 \\
Allstate & 256 & 2 & 0.10 \\
Bike & 108 & 2-3 & 0.15 \\
Boston & 32 & 3 & 0.30 \\
California & 64 & 2 & 0.10 \\
Parkinson's & 64 & 2 & 0.10 \\
Yelp & 512 & 3 & 0.15 \\
\bottomrule
\end{tabular}
\end{table}

\subsection{Regression Models}
To evaluate representation robustness across different learning paradigms, we use ridge regression~\cite{hoerl1970ridge}, XGBoost~\cite{chen2016xgboost}, and deep neural networks~\cite{goodfellow2016deep}.

Ridge regression provides a strong linear baseline for compressed representations. XGBoost is evaluated using the standard regression configuration. Neural network experiments use the \texttt{MLPRegressor} implementation from scikit-learn with two hidden layers (128 and 64 units), ReLU activation, Adam optimization, and a maximum of 500 training iterations.

Using models with different inductive biases allows evaluation of whether Bloom Filter encodings preserve predictive structure across both linear and nonlinear learners.

\subsection{Baseline Feature Representations}
Bloom Filter representations are compared against commonly used feature representations:
\begin{itemize}
\item \textbf{Raw Features:}  
Original numeric or tokenized features without compression.

\item \textbf{Principal Component Analysis (PCA)} \cite{jolliffe2002pca}:  
A linear dimensionality reduction technique that projects data onto orthogonal components maximizing explained variance.

\item \textbf{Random Projection (RP)} \cite{johnson1984extensions}:  
A randomized dimensionality reduction method that approximately preserves pairwise distances between samples.
\end{itemize}

To enable fair comparisons under memory constraints, PCA and RP dimensionalities were selected to produce encoded representations with approximately matched storage costs relative to the corresponding Bloom Filter configurations.

\subsection{Bloom Filter Structural Experiments}
Bloom Filter encodings are governed by structural parameters including filter size $m$ and the number of hash functions $k$. To characterize how these parameters influence predictive fidelity and representation sparsity, we conduct parameter sweeps over Bloom Filter size and hash count using the California Housing dataset.

Additional experiments compare multiple hash families to evaluate the sensitivity of Bloom Filter encodings to hash function choice. The evaluated hash functions include MurmurHash3, xxHash, FNV-1a, CRC32, MD5, SHA-1, SHA-256, and BLAKE2, representing a mix of non-cryptographic and cryptographic hash functions commonly used in practice \cite{appleby2011murmurhash,jenkins1997hash,neves2015blake2}.

\subsection{Entropy-Based Puncturing Experiments}
To evaluate entropy-based structural compression, we perform threshold sweeps over the entropy parameter $\tau$ using the Parkinson’s Telemonitoring dataset. For each threshold value, bit positions with entropy below $\tau$ are removed from the Bloom Filter representation.

Because the entropy of a Bernoulli variable is maximized when the activation probability approaches $0.5$, increasing the entropy threshold progressively removes bit positions whose activation probability deviates strongly from this balanced regime.

In addition to the primary low-entropy pruning strategy, comparative experiments remove high-entropy bits and randomly selected bits to further examine how predictive signal is distributed across the Bloom Filter representation.

\subsection{Structural Robustness Experiments}
To further analyze the structural properties of Bloom Filter encodings, additional experiments examine how predictive performance responds to controlled transformations of the encoded bit vectors.

\paragraph{Permutation Experiments}
Bloom Filter bit positions have no intrinsic semantic ordering because features are mapped through hash functions. We therefore perform permutation experiments in which Bloom Filter bit vectors are randomly shuffled prior to model training. Stable performance under permutation indicates that predictive signal is encoded in aggregate activation patterns rather than specific bit locations.

\paragraph{Bloom Filter Folding}
We additionally evaluate Bloom Filter folding as a structural compression mechanism. Folding compresses the representation by combining segments of the Bloom Filter using bitwise OR operations, reducing the number of stored bits while increasing collision rates through superposition of bit positions.

\subsection{Evaluation Metrics}
\label{sec:metrics}
Model performance is evaluated using predictive fidelity, representation size, and a storage-normalized summary of their trade-off. Predictive fidelity is assessed using the coefficient of determination ($R^2$), which measures the proportion of target variance explained by a model.

Representation cost is measured as the number of bytes required to store one encoded sample. Bloom Filter and punctured Bloom Filter representations are counted as packed binary arrays, whereas numeric representations include the bytes required by their stored components. This per-sample measure excludes shared preprocessing transformations and model parameters because the analysis focuses on the storage or transmission cost of individual encoded samples.

To summarize predictive performance relative to representation size, we define \emph{predictive efficiency} (PE) as
\begin{equation}
\mathrm{PE}=\frac{R^2}{B},
\end{equation}
where $B$ is the encoded representation size in bytes per sample. PE is therefore a storage-normalized summary statistic with units of inverse bytes; it is not intended as an information-theoretic measure of predictive information.

Because a ratio of this form can favor extremely small representations despite substantial predictive degradation, PE is interpreted jointly with $R^2$ rather than as a standalone performance criterion. In the PBF efficiency comparisons, configurations are considered only when their $R^2$ reaches at least 90\% of the best observed predictive performance for the corresponding dataset. This fidelity requirement prevents nearly empty representations from receiving favorable rankings solely because of their small denominator. PE is used to compare representations within each dataset under memory constraints, rather than as a universal measure for comparison across unrelated regression tasks.

%%%%%%%%%%%%%%%%%%%%%%%%%%%

\section{Results}
\label{sec:results}
\subsection{Evaluation Perspective}
Results are interpreted using the complementary measures defined in Section~\ref{sec:metrics}. Predictive fidelity is assessed using $R^2$, while predictive efficiency summarizes fidelity relative to per-sample representation size. Because marginal bit entropy measures variability rather than target relevance, the experiments empirically evaluate whether removing near-constant Bloom Filter positions reduces storage while preserving predictive fidelity.

\subsection{Best Raw vs. Best Compressed vs. Punctured Bloom Filters}
To contextualize predictive fidelity across representations, we compare:
\begin{enumerate}
\item The best-performing model trained on raw features.
\item The best-performing compressed baseline (PCA or Random Projection) under an approximately matched storage budget.
\item The best-performing Punctured Bloom Filter (PBF) model.
\end{enumerate}

Table~\ref{tab:best_r2_comparison} summarizes this comparison across datasets.

\begin{table}[t]
\centering
\caption{Best predictive performance ($R^2$) achieved for each representation type across datasets.}
\label{tab:best_r2_comparison}
\small
\begin{tabular}{lccc}
\toprule
Dataset & Raw & Non-BF & PBF \\
\midrule
Abalone    & 0.593 & 0.627 & 0.591 \\
Airline    & 0.994 & 0.990 & 0.879 \\
Allstate   & 0.567 & 0.521 & 0.558 \\
Bike       & 0.943 & 0.938 & 0.912 \\
Boston     & 0.882 & 0.785 & 0.819 \\
California & 0.852 & 0.795 & 0.764 \\
Parkinson  & 0.908 & 0.759 & 0.868 \\
Yelp       & 0.669 & 0.625 & 0.675 \\
\bottomrule
\end{tabular}
\end{table}

Across datasets, Punctured Bloom Filters maintain competitive predictive performance relative to both raw features and classical dimensionality reduction methods despite substantially reduced representation sizes. These results indicate that Bloom Filter encodings preserve sufficient predictive structure from the original feature space while operating under constrained storage budgets.

\subsection{Predictive Efficiency: \texorpdfstring{$R^2$}{R2} Relative to Encoded Representation Size}
Because all compression baselines were configured to operate under approximately matched storage budgets, predictive fidelity alone is insufficient to fully characterize representation quality. Different representations may achieve similar predictive fidelity while retaining substantially different amounts of predictive signal relative to encoded representation size. We therefore analyze predictive efficiency, as defined in Section~\ref{sec:metrics}, to evaluate how effectively each representation preserves predictive performance under constrained-memory settings.

Table~\ref{tab:r2_efficiency} reports predictive efficiency across datasets. To avoid degenerate efficiency values caused by extremely aggressive compression (e.g., puncturing nearly all bits), PBF configurations were restricted to those achieving at least 90\% of the best observed predictive performance for the corresponding dataset. This threshold serves as a practical guideline to balance predictive fidelity and compression efficiency.

\begin{table}[t]
\centering
\caption{Predictive efficiency measured as $R^2$ relative to encoded representation size per sample.}
\label{tab:r2_efficiency}
\small
\begin{tabular}{lccc}
\toprule
Dataset & Non-BF & PBF & Best \\
\midrule
Abalone    & 0.0203 & \textbf{0.0293} & PBF \\
Airline    & 0.0309 & \textbf{0.0898} & PBF \\
Allstate   & 0.0027 & \textbf{0.0093} & PBF \\
Bike       & \textbf{0.0085} & 0.0084 & RP \\
Boston     & 0.0082 & \textbf{0.0380} & PBF \\
California & 0.0124 & \textbf{0.0449} & PBF \\
Parkinson's& 0.0105 & \textbf{0.0720} & PBF \\
Yelp       & 0.0033 & \textbf{0.0236} & PBF \\
\bottomrule
\end{tabular}
\end{table}

In seven of the eight datasets, the Punctured Bloom Filter configuration achieves the highest predictive efficiency. Even in the single dataset where PBF is not ranked first, it remains competitive under similar storage budgets.

These results reinforce the central objective of this work: preserving predictive fidelity while reducing encoded representation size, with PE used to summarize the resulting trade-off.

\subsection{Bloom Filter Parameter Sensitivity Analysis}
To evaluate how Bloom Filter structural parameters affect predictive fidelity, we perform a parameter sweep over filter size ($m$) and number of hash functions ($k$) using the California Housing dataset. This dataset was selected because of its moderate size and stable baseline performance, making it suitable for controlled parameter analysis.

\begin{figure}
\centering
\includegraphics[width=0.975\columnwidth]{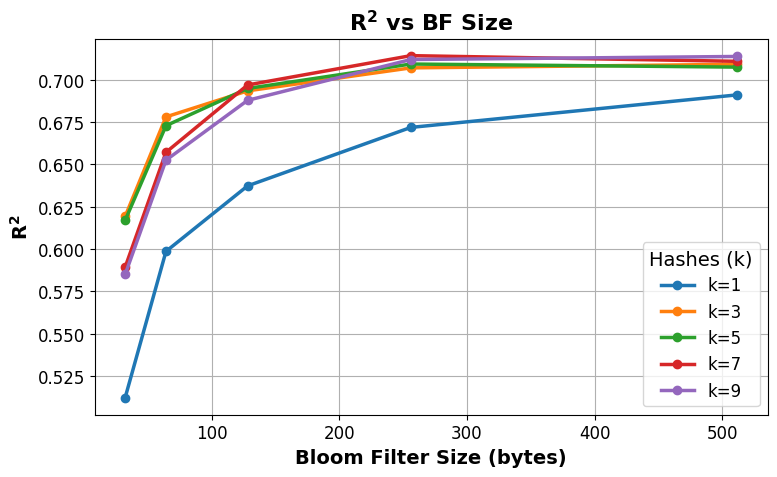}
\caption{Effect of varying Bloom Filter size ($m$) and hash count ($k$) on XGBoost $R^2$ performance for the California Housing dataset.}
\label{fig:bf_sweep_r2}
\vspace{-2mm}
\end{figure}

\begin{figure}
\centering
\includegraphics[width=0.975\columnwidth]{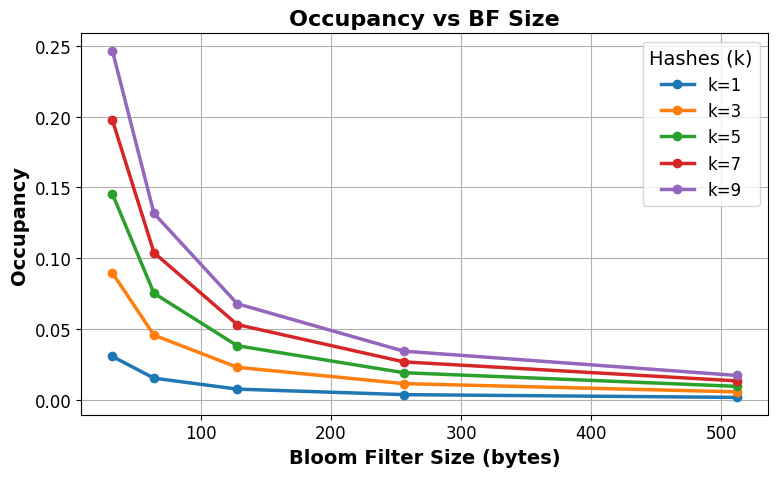}
\caption{Bit occupancy as a function of Bloom Filter size and hash count.}
\label{fig:bf_sweep_bo}
\vspace{-2mm}
\end{figure}

\begin{figure}
\centering
\includegraphics[width=0.975\columnwidth]{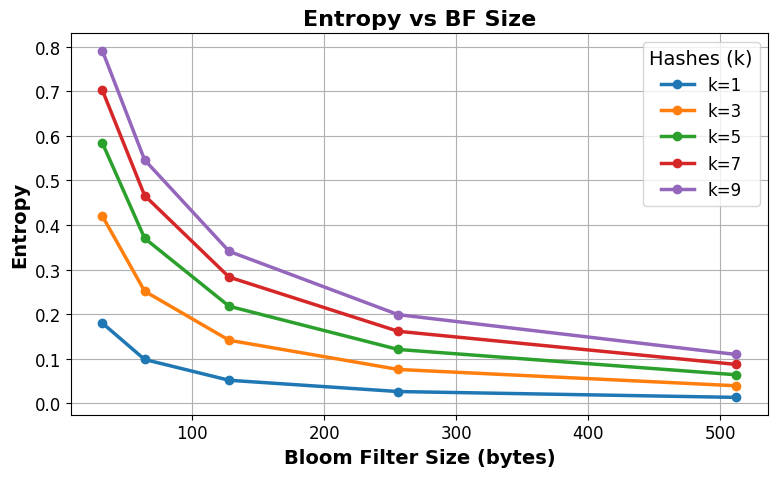}
\caption{Entropy behavior under varying Bloom Filter size and hash count.}
\label{fig:bf_sweep_entropy}
\vspace{-2mm}
\end{figure}

Figures~\ref{fig:bf_sweep_r2}-\ref{fig:bf_sweep_entropy} illustrate the effect of varying Bloom Filter size and hash count on predictive fidelity, bit occupancy, and entropy behavior.

Across all experiments, larger Bloom Filters combined with moderate hash counts consistently provide the best trade-off between predictive fidelity and representation sparsity. Increasing filter size reduces token collisions, while moderate hash counts maintain adequate bit dispersion without introducing excessive redundancy.

These results confirm that Bloom Filter structural parameters materially influence predictive fidelity and representation behavior.

\subsection{Impact of Entropy Threshold on Predictive Efficiency}
This experiment evaluates whether low-entropy Bloom Filter positions can be removed to improve representation efficiency while preserving predictive fidelity. To analyze the effect of structural compression, we perform an entropy-threshold sweep using the Parkinson's Telemonitoring dataset. For each threshold $\tau$, bit positions whose entropy falls below $\tau$ are removed from the Bloom Filter representation.

\begin{figure}
\centering
\includegraphics[width=0.975\columnwidth]{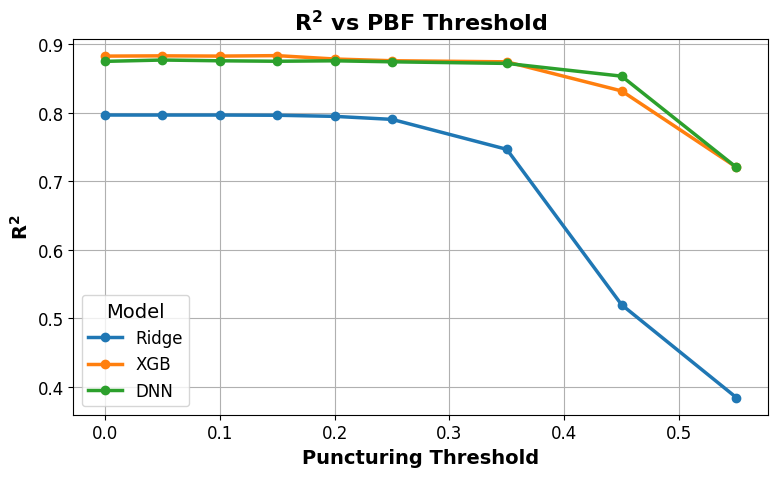}
\caption{Predictive fidelity ($R^2$) as a function of entropy puncturing threshold for the Parkinson’s dataset.}
\label{fig:pbf_threshold_r2}
\vspace{-2mm}
\end{figure}

\begin{figure}
\centering
\includegraphics[width=0.975\columnwidth]{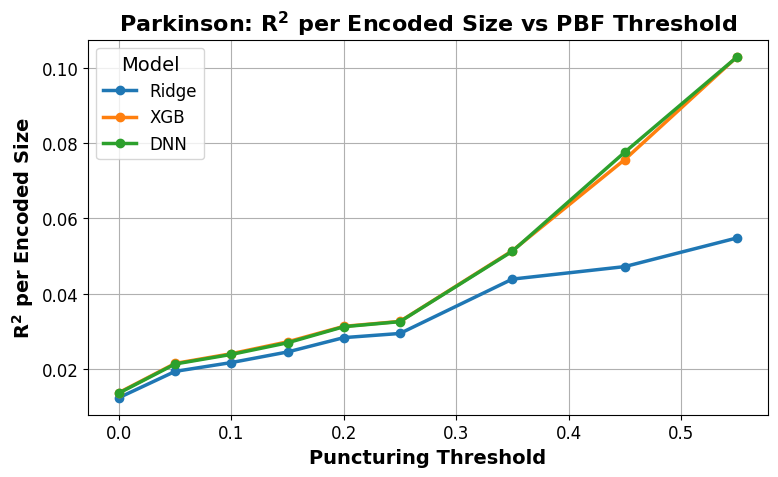}
\caption{Predictive efficiency as a function of entropy puncturing threshold for the Parkinson’s dataset.}
\label{fig:pbf_threshold_r2byte}
\vspace{-2mm}
\end{figure}

Figure~\ref{fig:pbf_threshold_r2} shows predictive fidelity as a function of the entropy threshold, while Figure~\ref{fig:pbf_threshold_r2byte} illustrates predictive efficiency. At low thresholds, puncturing removes highly saturated or inactive bits, resulting in substantial reductions in representation size with minimal loss in predictive fidelity. As the threshold increases, progressively more bits are removed, eventually reducing the representational capacity of the Bloom Filter.

The results show that moderate entropy thresholds provide the most favorable efficiency trade-off. In particular, Figure~\ref{fig:pbf_threshold_r2} shows that thresholds between approximately 0.15 and 0.35 remove a large fraction of low-variability positions while maintaining nearly identical predictive fidelity. Across cross-validation folds, predictive fidelity remained stable under moderate entropy thresholds. For example, using XGBoost on the Parkinson's dataset, the configuration with $\tau = 0.15$ achieved $R^2 = 0.8830 \pm 0.0083$ across folds while reducing the representation size from 64 bytes to approximately 32 bytes per sample.

As the threshold increases toward maximum binary entropy, positions with progressively greater marginal variability are removed, and predictive fidelity begins to degrade. Overall, the threshold sweep demonstrates that entropy-based puncturing provides an effective and computationally inexpensive mechanism for structural compression.

\subsection{Structural Robustness Experiments}
To further analyze how predictive signal is distributed within Bloom Filter representations, we conducted additional structural experiments examining controlled bit removal and representation folding.

\begin{figure}
\centering
\includegraphics[width=0.975\columnwidth]{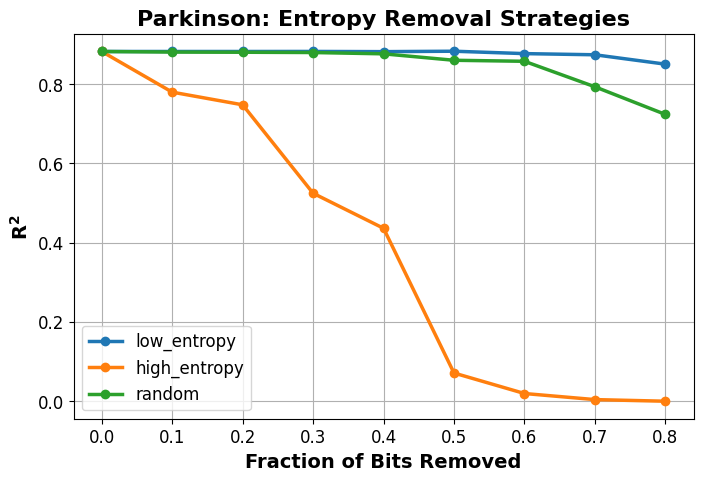}
\caption{Predictive fidelity ($R^2$) as a function of the fraction of Bloom Filter bits removed under different removal strategies.}
\label{fig:entropy_r2}
\vspace{-2mm}
\end{figure}

\paragraph{Entropy-Guided Bit Removal}
Figure~\ref{fig:entropy_r2} shows predictive fidelity as a function of the fraction of Bloom Filter bits removed under three strategies: removing low-entropy bits, removing high-entropy bits, and random removal.

Removing low-entropy bits preserves predictive fidelity even when a substantial portion of the representation is removed. In contrast, removing high-entropy bits rapidly degrades predictive fidelity, while random removal exhibits intermediate behavior. For the evaluated representation, removing higher-entropy positions is more damaging than removing lower-entropy or randomly selected positions. This result supports entropy as a practical ranking criterion for puncturing in this setting, without implying that entropy directly measures target relevance.

\begin{figure}
\centering
\includegraphics[width=0.975\columnwidth]{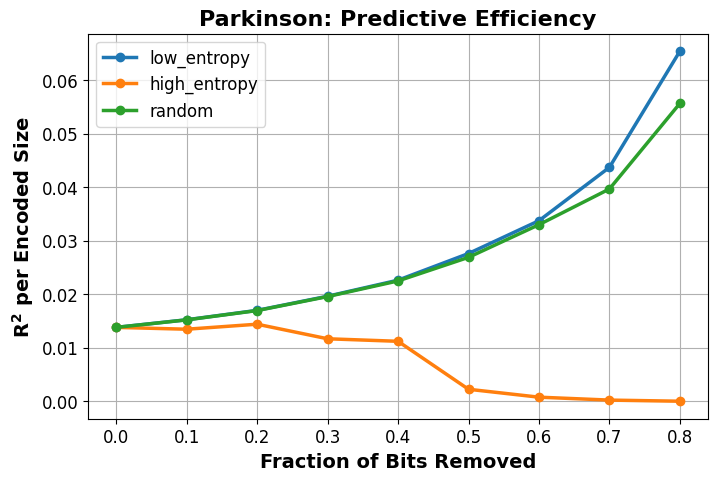}
\caption{Predictive efficiency under different Bloom Filter bit removal strategies.}
\label{fig:entropy_eff}
\vspace{-2mm}
\end{figure}

Figure~\ref{fig:entropy_eff} further illustrates predictive efficiency under the different removal strategies. Under low-entropy removal, predictive efficiency increases over moderate removal levels as representation size decreases while predictive fidelity remains relatively stable. This behavior reflects the storage-performance trade-off summarized by PE.

\paragraph{Bloom Filter Folding}
To further examine structural redundancy in Bloom Filter representations, we evaluated OR-based Bloom Filter folding on the Parkinson’s dataset. Folding compresses the representation by combining bit positions using logical OR operations, reducing representation size while increasing collision rates through bit superposition.

Table~\ref{tab:folding_results} summarizes predictive fidelity and predictive efficiency across multiple fold ratios using XGBoost regression. Results show that moderate folding introduces only minimal degradation in predictive fidelity despite substantial reductions in representation size. For example, reducing the representation from 64 bytes to 32 bytes preserves nearly identical predictive performance while substantially improving predictive efficiency.

These results suggest that Bloom Filter representations contain substantial structural redundancy that can be exploited for compression. However, unlike entropy-guided puncturing, folding increases collisions indiscriminately and therefore may eventually degrade predictive fidelity under more aggressive compression.

\begin{table}[b]
\centering
\caption{Effect of Bloom Filter folding on predictive fidelity and predictive efficiency for the Parkinson's dataset using XGBoost.}
\label{tab:folding_results}
\small
\begin{tabular}{lccc}
\toprule
Keep Ratio & Bytes & $R^2$ & $R^2$/Byte \\
\midrule
1.00 & 64 & 0.8824 & 0.0138 \\
0.90 & 58 & 0.8828 & 0.0152 \\
0.80 & 52 & 0.8819 & 0.0170 \\
0.70 & 45 & 0.8809 & 0.0196 \\
0.60 & 39 & 0.8752 & 0.0224 \\
0.50 & 32 & 0.8807 & 0.0275 \\
\bottomrule
\end{tabular}
\end{table}

\subsection{Storage-Matched Feature Hashing Comparison}
To determine whether the observed performance results simply from mapping quantized tokens into a compact hashed representation, we conducted a focused comparison with standard signed feature hashing~\cite{weinberger2009feature}. Feature hashing used the same quantized feature-value tokens, cross-validation splits, and XGBoost configuration as PBF. Hash dimensions were fixed before evaluation to approximately match PBF storage: six \texttt{float32} values for California Housing and eight for Parkinson's Telemonitoring. The PBF operating points used here were selected for storage matching and therefore need not coincide with the maximum-efficiency PBF configurations reported in Table~\ref{tab:r2_efficiency}.

\begin{table}[t]
\centering
\caption{Storage-matched comparison with feature hashing.}
\label{tab:feature_hashing}
\small
\setlength{\tabcolsep}{3.5pt}
\begin{tabular}{llccc}
\toprule
Dataset & Method & $R^2$ & Bytes & PE \\
\midrule
California
  & PBF & \textbf{0.7622} & 23.6 & \textbf{0.0323} \\
  & FH  & 0.0844 & 24.0 & 0.0035 \\
Parkinson's
  & PBF & \textbf{0.8830} & 32.4 & \textbf{0.0273} \\
  & FH  & 0.3654 & 32.0 & 0.0114 \\
\bottomrule
\end{tabular}
\end{table}

As shown in Table~\ref{tab:feature_hashing}, PBF retained substantially greater predictive fidelity and predictive efficiency at nearly identical storage budgets. The feature-hashing results were also consistent across folds, with $R^2=0.0841\pm0.0052$ for California and $R^2=0.3654\pm0.0153$ for Parkinson's. Although this focused experiment does not establish general superiority over feature hashing, it demonstrates that the performance observed in these settings cannot be attributed solely to fixed-length hashing or collision-based compression.

%%%%%%%%%%%%%%%%%%%%%%%%%%%

\section{Discussion}
The results demonstrate that Bloom Filter encodings can provide compact representations for regression while preserving useful predictive fidelity across diverse datasets. Entropy-punctured Bloom Filters were competitive with classical dimensionality-reduction methods in many settings and generally improved predictive efficiency relative to encoded representation size. In a storage-matched comparison on California Housing and Parkinson's Telemonitoring, PBF also retained substantially greater predictive fidelity than standard signed feature hashing. Although this focused comparison does not establish general superiority over feature hashing, it indicates that the observed PBF performance cannot be attributed solely to fixed-length hashing or collision-based compression.

The proposed method follows an expansion-compression paradigm. Initial Bloom Filter encoding introduces structured redundancy through $k$-hashing and token superposition, after which entropy-based puncturing removes low-variability bit positions. The structural experiments show that removing lower-entropy positions is less damaging than removing higher-entropy or randomly selected positions over moderate compression levels. This empirical asymmetry enables substantial representation reduction with limited degradation in predictive fidelity, but it should not be interpreted as evidence that entropy directly measures target relevance.

The results also highlight the importance of evaluating feature representations under explicit memory constraints. Predictive fidelity alone does not describe the trade-off between model performance and representation cost. Predictive efficiency therefore provides a complementary storage-normalized summary when compact representations are required. It should be interpreted as a descriptive, within-dataset measure rather than as an information-theoretic quantity or a replacement for $R^2$.

The structural robustness experiments further indicate that Bloom Filter representations contain substantial redundancy. Moderate entropy-guided puncturing removes low-variability components while preserving predictive structure, whereas more aggressive transformations, such as folding, increase collisions and eventually degrade predictive fidelity.

Several limitations remain. Bloom Filter encodings inherently introduce collisions that can obscure fine-grained feature distinctions, particularly under aggressive compression. Effective parameter choices also depend on dataset characteristics, representation sparsity, and tokenization strategy. Although the lightweight heuristic used in this work produced practical operating regimes across the evaluated datasets, it does not guarantee globally optimal configurations. Moreover, entropy is a marginal, unsupervised measure of bit variability rather than a measure of mutual information with the regression target. Low-entropy positions may therefore remain predictive, and entropy-based puncturing may remove useful information in some settings.

Future work may investigate supervised bit-selection criteria, adaptive parameter selection, alternative tokenization strategies, and applications in distributed or privacy-sensitive machine learning settings where compact hashed feature representations are beneficial.

%%%%%%%%%%%%%%%%%%%%%%%%%%%

\section{Conclusion}
This work investigated Bloom Filter encodings as compact feature representations for memory-efficient regression and introduced entropy-based puncturing as a representation-level compression mechanism. By removing low-variability bit positions using statistics estimated from the training data, the proposed method reduces encoded representation size while preserving predictive fidelity.

Experiments across diverse regression datasets showed that Bloom Filter encodings can remain competitive with classical dimensionality-reduction methods, while entropy-guided puncturing generally improves predictive efficiency relative to encoded representation size. In focused storage-matched experiments, PBF also retained substantially greater predictive fidelity than standard signed feature hashing on California Housing and Parkinson's Telemonitoring. The structural experiments further demonstrated that moderate entropy-guided puncturing can remove substantial representational redundancy with limited performance degradation, while more aggressive compression eventually reduces predictive fidelity.

Overall, entropy-punctured Bloom Filters provide a practical approach for combining fixed-length feature encoding with structural compression under memory constraints. Future work will investigate supervised bit-selection criteria, adaptive parameter selection, alternative tokenization strategies, and distributed or privacy-sensitive learning settings in which compact hashed representations may reduce storage or communication cost.

\bibliographystyle{IEEEtran}
\bibliography{conf_paper_references}

\vspace{12pt}

\end{document}